\documentclass[runningheads]{llncs}
\usepackage{graphicx}    
\usepackage{epstopdf}
\usepackage{color}
\usepackage{cite}
\usepackage{amsfonts,amssymb,amsmath}
\usepackage{multirow}
\usepackage{epsf}
\usepackage{longtable}
\usepackage{booktabs}
\usepackage[table]{xcolor}
\usepackage{ragged2e}
\usepackage{subfigure}
\usepackage{marvosym}
\usepackage{rotating}
\usepackage{url}
\begin{document} 
%
% \title{Multi-view Classification Network with Impartial Decision for Frontotemporal Dementia }
\title{A Multi-view Impartial Decision Network for Frontotemporal Dementia Diagnosis}
\author{Guoyao Deng\inst{1}, Ke Zou\inst{1,3}, Meng Wang\inst{3}, Xuedong Yuan\inst{2}, Sancong Ying\inst{2} and Huazhu Fu\inst{3}} 
\authorrunning{G.Deng et al.}
\institute{National Key Laboratory of Fundamental Science on Synthetic Vision,\\ Sichuan
University, Sichuan, China \and College of Computer Science, Sichuan University, Sichuan, China \and  Institute of High Performance Computing, A*STAR, Singapore
}
%\thanks{Supported by Sichuan Key R\&D Program (Grant No. 2020YFG0075) }}
% my \orcidID{0000-0002-2025-9859}
% \author{Anonymous}
% %
% \authorrunning{Anonymous et al.}
% \institute{Anonymous Organization\\
% %\email{yxd@scu.edu.cn} 
% \email{**@******.***}
% }
%
\maketitle              % typeset the header of the contribution
\begin{abstract}
Frontotemporal Dementia (FTD) diagnosis has been successfully progress using deep learning techniques. However, current FTD identification methods suffer from two limitations. Firstly, they do not exploit the potential of multi-view functional magnetic resonance imaging (fMRI) for classifying FTD. Secondly, they do not consider the reliability of the multi-view FTD diagnosis.
To address these limitations, we propose a reliable multi-view impartial decision network (MID-Net) for FTD diagnosis in fMRI. Our MID-Net provides confidence for each view and generates a reliable prediction without any conflict. To achieve this, we employ multiple expert models to extract evidence from the abundant neural network information contained in fMRI images. We then introduce the Dirichlet Distribution to characterize the expert class probability distribution from an evidence level. Additionally, a novel Impartial Decision Maker (IDer) is proposed to combine the different opinions inductively to arrive at an unbiased prediction without additional computation cost.
Overall, our MID-Net dynamically integrates the decisions of different experts on FTD disease, especially when dealing with multi-view high-conflict cases. Extensive experiments on a high-quality FTD fMRI dataset demonstrate that our model outperforms previous methods and provides high uncertainty for hard-to-classify examples. We believe that our approach represents a significant step toward the deployment of reliable FTD decision-making under multi-expert conditions. We will release the codes for reproduction after acceptance.
% rather than segmentation robustness and reliable uncertainty estimations
\end{abstract}
\section{Introduction}
Frontotemporal Dementia (FTD) has become the second most common type of presenile dementia, which causes personality changes, progressive behavioral problems, and cognitive impairment in tens of millions of the elderly. FTD includes a range of subtypes. This heterogeneity of FTD hampers the diagnosis process. Thus early and accurate diagnoses are of vital importance for comprehending the disease process and developing disease-modifying treatments \cite{bright2019neuroinflammation,meeter2017imaging}. For deep learning computer-aid-diagnosis based on fMRI images, since fMRI images inherently have multiple perspectives on brain neural network activity, multi-view input can enable the network to learn abundant brain activity information. But two major problems exist in current FTD deep learning classification: Single-view classification fails to extract sufficient brain activity information. And fusion strategies in trusted multi-view classification do not properly balance conflicts.

There are a variety of CAD methods currently in fMRI image diagnosis~\cite{rashid2020use,zhang2021survey}. Traditional machine learning CAD methods require hand-crafted features and feature selection, which ignore abundant origin information hidden in fMRI images~\cite{pereira2009machine}. Deep learning CAD methods using networks like CNN  mostly focus on directly extracting information from fMRI images without feature egineering~\cite{10.3389/fninf.2018.00023,mao2019spatio}. Sarraf \textit{et al.}~\cite{sarrafclassification} used LeNet-5 to classify Alzheimer’s Disease (AD) from healthy controls using single-view 2D images converted from fMRI data. Ramzan \textit{et al.}~\cite{ramzan2020deep} applied Resnet to classify AD with the same input data form. These approaches can not correctly estimate the confidence of the predictions because of the drawback of softmax. Moreover, the promise of multi-view classification in fMRI images has not been investigated. To this end, a reliable model should both recognize sufficient brain neural network information and provide accurate confidence.

Multi-view classification is mainly divided into early fusion, late fusion, and score fusion according to fusion strategy\cite{do2017plant,zhang2018multi}.
Confidence-based fusion proposed by Han \textit{et al.} \cite{hantrusted} is a late fusion strategy that combines decisions at the evidence level. It combined multi-view information with Dempster-Shafer Theory aiming to provide trusted prediction with confidence. Although this method did promote both classification reliability and robustness, the fatal "Zadeh's problem" remained unsolved \cite{zadeh1984review}. While trying to combine two opinions that conflict with each other using DS-Combine\cite{shafer1976mathematical}, the combined result becomes counter-intuitive and baseless. This reveals that DS-Combine is inapplicable in FTD diagnosis due to such results may affect the doctor's judgment, and even delay the diagnosis and effective treatment of FTD disease\cite{1255332}. A balanced and risk-aware method is needed.

Based on the above analysis, we propose a credible Frontotemporal Dementia (FTD) multi-view classification method in this paper. Our model infers confidence for each view's prediction and properly handles the complementary and conflicting information among them, providing a more accurate and reliable diagnosis of FTD. To estimate the confidence of each view, we adopted evidential deep learning. Furthermore, we proposed the Impartial Decision Maker (IDer) to solve "Zadeh's problem" in current trusted multi-view classification, avoiding high-risk and counter-intuitive prediction. This approach is a crucial step towards safe and trusted FTD diagnosis and deployment.

To the best of our knowledge, we are the first to address conflict in FTD multi-view classification. Our contributions can be summarized as follows:\\
(1) We propose a novel multi-view impartial decision method (MID-Net) for Frontotemporal Dementia diagnosis and classification based on rs-fMRI images.\\
(2) We introduce the Impartial Decision Maker (IDer) for sample-adaptive multi-view integration, which combines multi-view information at the evidence level and forms a unified opinion without any conflict.\\
(3) We conduct sufficient experiments on the FTD dataset\footnote{The dataset has IRB certification.} to verify the performance of our MID-Net and the effectiveness of uncertainty estimation.

\section{Method}
 %In neurodegenerative diseases, structural abnormalities are often preceded by functional changes. Resting-state functional MRI is an MRI technique that measures functional connectivity between brain regions to differentiate pathological sub-types of FTD.\\
%On the one hand, using softmax for classification often results in the model outputting a high degree of confidence even for incorrect predictions. On the other hand, for FTD fMRI image classification, making use of the complementary information that exists in multiple brain regions helps improve the reliability and accuracy of predictions. Multi-view evidential deep learning is the most proper way to solve the above problems. Therefore we proposed a multi-view classification method with a balanced evidential fusion strategy for reliable and risk-aware decision making.

 \begin{figure}[!t]
    \centering
    \includegraphics[width=12cm]{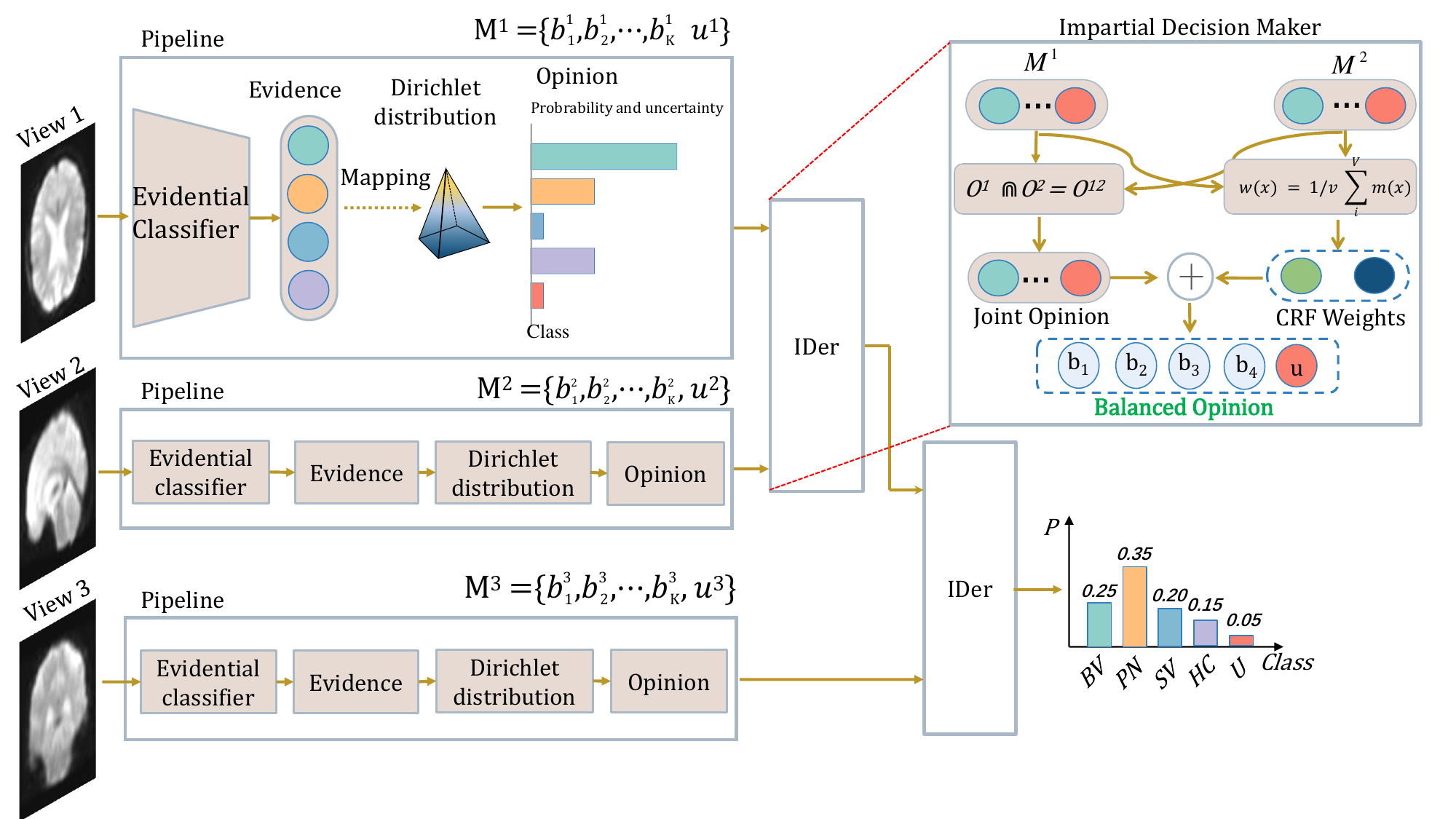}
    \caption{Framework of our MID-Net. Each pipeline analyzes different views and forms its own opinion, and the IDer is used to integrate these individual opinions.
Opinion}
    \label{fig:framework}
\end{figure}

\subsection{Overall framework \& Uncertainty quantification}
The overall framework is shown in Fig.~\ref{fig:framework}. In order to make full use of brain activity information in rs-fMRI images, we first deploy backbone networks for three sectional views. After the feature extraction, models form the opinions with collected evidence through a Dirichlet distribution. At the final decision level, we utilize IDer to complete the trusted fusion. We now delve into the details.

\noindent \textbf{Evidential Classifier:} 
The drawback of the maximum-likelihood classifier has been discussed, that the point estimate of softmax produces over-confident erroneous prediction when faces out-of-distribution samples. When deployed in CAD, it will cause ineffective treatment and even irreparable consequences. In order to quantify the confidence behind every choice our model makes for a single view, we utilize evidential classifiers for more reliable predictions, which infer the strength of evidence to quantify belief masses and uncertainty under Dirichlet distribution\cite{sensoy2018evidential}. 

\noindent  \textbf{Dirichlet Distribution:}
In the theory of Subjective Logic \cite{josang2016subjective}, Dirichlet distribution formalizes the model's opinions assigned belief mass of any subset of the frame of discernment and the uncertainty mass as an explicit parameter. This type of uncertainty mass expresses "I do not know" for all possible states. More specifically, in a frame of K mutually exclusive singletons (e.g., class labels), these K belief masses and the uncertainty mass of each view are all non-negative and sum to one, as:
\begin{equation}
u_{}^v + \sum_{k=1}^K{b_k^v} = 1,
\label{E_1} 
\end{equation}
where $u_{}^v\geq0$ and $b_k\geq0$ for $k = 1,2,···, K$, denote the overall uncertainty and the belief of $k$-th class respectively. 
the evidence ${e_{}^v = [e_{1}^v,\cdots,e_{K}^v]}$ induce the parameter ${\alpha_k^v}$ of Dirichlet distribution in the theory of subjective theory,i.e., ${\alpha_k^v} = e_k^v + 1$. And the belief mass $b_k^v$ and uncertainty mass $u_{}^v$ are computed as:
\begin{equation}
b_k^v = \frac{e_k^v + 1}{S_{}v} = \frac{\alpha_v}{S_{}^v} \text{ and }  u_{}^v = \frac{K}{S_{}^v},  
\label{E_2}
\end{equation}
where $S_{}^v =\sum\limits_k^K\alpha_k^v = \sum\limits_k^K(e_k^v+1)$ refers to as the Dirichlet strength. For each view, the more evidence observed for a possible class from a sample, the greater the corresponding belief. When little evidence is gained, greater uncertainty is assigned to this view. Composing opinions via Dirichlet distribution avoids making false predictions that are overconfident, also the estimated uncertainty ensures the decision risk of our model is visible.

% This property is critical for CAD since we exactly know the reliability of the model's prediction regardless of whether the input sample is hard to classify, instead of leaving us guessing how sure is the result.
\subsection{Impartial Decision Maker}
%With sufficient brain activity information collected, we need an effective and efficient late fusion strategy to fuse them. the feature of score fusion that blindly treats all opinions equally make it directing fusing softmax output leads to over confidence even for erroneous predictions. 

\begin{figure}[!t]
    \centering
    \includegraphics[width=11cm]{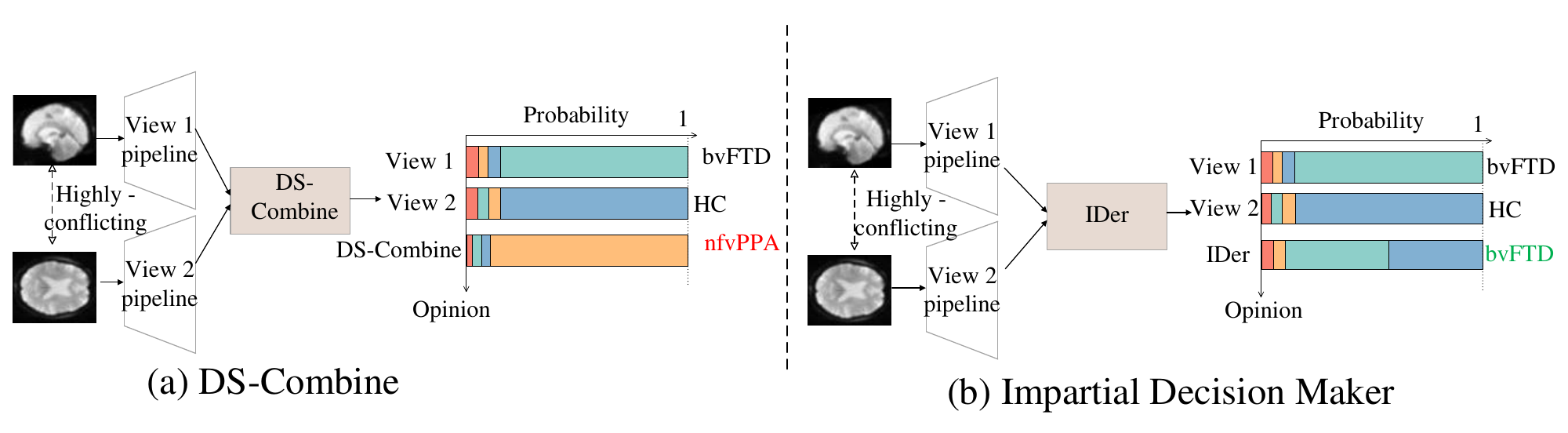}
    \vskip -5pt
    \caption{A example of FTD subject lead DS-combine\cite{hantrusted} to counterintuitive result while IDer predicts reasonably. }
    \label{fig:sample}
\end{figure}

\noindent  \textbf{"Zadeh's problem" in FTD:} As shown in Fig.~\ref{fig:sample} (a), due to the characteristics of complementary brain activity information contained in different brain regions\cite{fan2016human}, the conclusions drawn from different perspectives may be divergent. The Current fusion strategy DS-Combine cannot fuse two divided opinions from two different perspectives, even arrive at a counterintuitive, baseless opinion. We found that this problem of DS-combine is caused by using Dempster's rule to fuse opinions. This drawback of Dempster's rule has been pointed out by Zadeh \textit{et al.}\cite{zadeh1984review,dezert2006proportional}. Therefore, in the multi-view classification of fMRI images, a fusion strategy that can properly handle conflicts of opinion is critical.

To better fuse opinion with low risk and resolve the highly conflicting situation, we proposed the Impartial Decision Maker based on The Weighted Operator Theory \cite{1255332}aiming to achieve a more balanced information fusion. For any two views, models' opinions towards $K$ classes $O_{}^1 = [b_1^1,b_2^1,...,b_k^1,u_{}^1]$ and $O_{}^2= [b_1^2,b_2^2,...,b_k^2,u_{}^2]$ is 
 combined in the following manner:
\begin{equation}
 O^{IDer} = O_{}^1 \doublecap O_{}^2,
 \label{E_3}
\end{equation}
where $\doublecap$ is the combination operator. The specific formulation in our IDer is:
\begin{equation}
b_k^{IDer} = b_k^1b_k^2 + b_k^1u_{}^2 + b_k^2u_{}^1 + w_{b_k}^{CRF},u^{IDer} = u_{}^1u_{}^2+ w_u^{CRF},
\label{E_4}
\end{equation}
where$w_{b_k}^{CRF}$ and $w_{u}^{CRF}$ are the conflict resolution factor(CRF). $b_k^{IDer}$ is the combined belief of $k$ class and $u^{IDer}$ is the combined uncertainty of two views. CRF measures conflict among beliefs and redistribute belief and uncertainty. $w_{b_k}^{CRF}$ and $w_{u}^{CRF}$ are calculated as:
\begin{equation}
w_{b_k}^{CRF} = 1/2(b_{k}^1+b_{k}^2), 
 w_u^{CRF} = 1/2(u_{}^1+u_{}^2).
\label{E_5}
\end{equation}
Based on the above formulations, we obtain the unified opinion $O^{IDer}$ for two views.  %Then according to Eq. \ref{E_6}, evidence of multi-view combination $e_{}^{IDer} = \{e_k\}_{k=1}^K$ and the parameter of Dirichlet distribution $\alpha_{}^{IDer} = \{\alpha_k\}_{k=1}^K$ is calculated in the following manner:
%\begin{equation} 
%e_k = b_k\times S\text{,}S = \frac{K}{u}\text{ %and }\alpha_k = e_k+1.
%\label{E_6}
%\end{equation}
Thus given $v$ views of data, we can first collect evidence from each view and then combine the opinions from different views by the following rule:
\begin{equation}
 O^{IDer} = O_{}^1 \doublecap O_{}^2 \cdots O_{}^v.
 \label{E_7}
\end{equation}

After obtaining the final opinion $O^{IDer} =[\{b_k^{IDer}\}_{k=1}^{K},u^{IDer}]$ for all views, the combined evidence for all possible classes $e^{IDer}$ is calculated according to Eq. \ref{E_2}.  Finally, the probability of all categories and the uncertainty of the overall decision are inferred through the above parameters.

On account of the method above, IDer combines evidence elegantly from different sources in a balanced way as shown in Fig. \ref{fig:sample}. Complementary information from highly conflicting opinions is reasonably taken into account. Compared to the DS-combine evidential fusion rule\cite{hantrusted}, one can recognize that our IDer is more reasonable for multi-view fMRI image information fusion since different view sections of the neural network can contain very complementary information. With integrated opinions inferring the output in a human-understandable fashion, IDer guarantees impartiality in decision-making and truthful causality exists in our model's diagnosis, instead of leaving us questioning why is the result.

\subsection{Learning process for FTD} 

For a given sample $i$, our model output the evidence of each class, which is represented as $e_i$.Furthermore, the corresponding parameter of Dirichlet distribution $\alpha_i$ is equal to $e_i + 1$. By calculating this parameter, we can obtain the final estimation$\frac{a_i}{S_i}$ of class probabilities. As our downstream task is a classification task, we first need cross-entropy loss function to supervise the training process, In the multinomial opinion $D(p_i|\alpha_i)$, where $p_i$ is the class probability. Therefore, we adopted the adjusted CE loss which can be further addressed as follows:
\begin{equation}
\small
\label{E_8}
{{\mathcal L}_{a}} = \int {\left[ {\sum\limits_{n = 1}^C { - {y_{ik}}\log ({p_{ik}})} } \right]} \frac{1}{{B\left( {{\alpha _i}} \right)}}\prod\limits_{n = 1}^C {{p_{ik}}^{{\alpha _{ik}} - 1}d{{{\bf{p}}_i}} = \sum\limits_{n = 1}^C {{y_{ik}}\left( {\psi \left( {{S_i}} \right) - \psi \left( {{\alpha _{ik}}} \right)} \right)}},
\end{equation}
where $\psi \left(  \cdot  \right)$ denote the $digamma$ function. ${{\bf{p}}_m}$ is the class assignment probabilities on a simplex. To guarantee that incorrect labels will yield less evidence, even shrinking to 0, the KL divergence loss function is introduced as below:
\begin{equation}
\small
{{\mathcal L}_{KL}} = \log \left( {\frac{{\Gamma \left( {\sum\nolimits_{j= 1}^K {{{\tilde \alpha }_{ij}}} } \right)}}{{\Gamma (K)\sum\nolimits_{j= 1}^C {\Gamma \left( {{{\tilde \alpha }_{ij}}} \right)} }}} \right) + \sum\nolimits_{j = 1}^K {\left( {{{\tilde \alpha }_{ij}} - 1} \right)} \left[ {\psi \left( {{{\tilde \alpha }_{ij}}} \right) - \psi \left( {\sum\nolimits_{j = 1}^C {{{\tilde \alpha }_{ij}}} } \right)} \right],
\label{E_9}
\end{equation}

where $\Gamma \left(  \cdot  \right)$ is the $gamma$ function. ${{\tilde \alpha }_{mc}} = {y_{mc}} + \left( {1 - {y_{mc}}} \right) \odot {\alpha _{mc}}$ denotes the adjusted parameters of the Dirichlet distribution, which is used to ensure that ground truth class evidence is not mistaken for 0. Hence, the overall loss function of our proposed network can be defined as follows:
\begin{equation}
\label{E_7}
{\mathcal L} = {{\mathcal L}_{a}} + {\lambda}{{\mathcal L}_{KL}},
\end{equation}
where ${\lambda}>0$  is the annealing factor, in order to prevent the training process from collapsing at the early stage thereby ending up with a flat uniform distribution output.

\section{Experiments}
 In order to evaluate the proposed method, we compared with the following methods: single-view softmax classifier (S-S), single-view evidential deep learning (S-E), multi-view softmax classifiers with score fusion (M-S+SF), multi-view evidential classifiers with DS-combine (M-E+DS). MLE methods are supervised by cross-entropy loss. Two backbone networks, Reset-18 and Vision Transformer are chosen. 
 
\textbf{Data \& Implementation Details} We validate our method on the FTD test set. All the data are pre-processed by SPM12 \footnote{https://www.fil.ion.ucl.ac.uk/spm/}and DPABI \cite{Yan_2016} and calculated in MNI space. 164 cases of patients with ground truth are stratified and divided into train, validation, and test sets. The test set size is 20\% of the total dataset size. The 4D fMRI image data in nifti format was first converted along the time axis into a stack of 3D volumes. Then we extract horizontal, lateral, and frontal view sections from the center of each volume. The shape of the input slice is zoomed to  $3\times224\times244$. The data contain 4 classes, which are labeled as bvFTD (label 0), svPPA(label 1), healthy control(label 2), and nfvPPA (label 4). Our proposed network is implemented in PyTorch and trained on NVIDIA GeForce RTX 3090. We adopt the Adam optimizer to optimize the overall parameters with an initial learning rate of 0.001. The maximum epoch is set to 10. Random hue change, random saturation change, random value change of the image, Gaussian blur, motion blur, and median blur are utilized as data augmentation. All the following experiments adopted a five-fold cross-validation strategy to prevent performance improvement caused by accidental factors.

\begin{figure}[!t]
\centering
\label{Fig.uncertainty}
\includegraphics[width = 12cm]{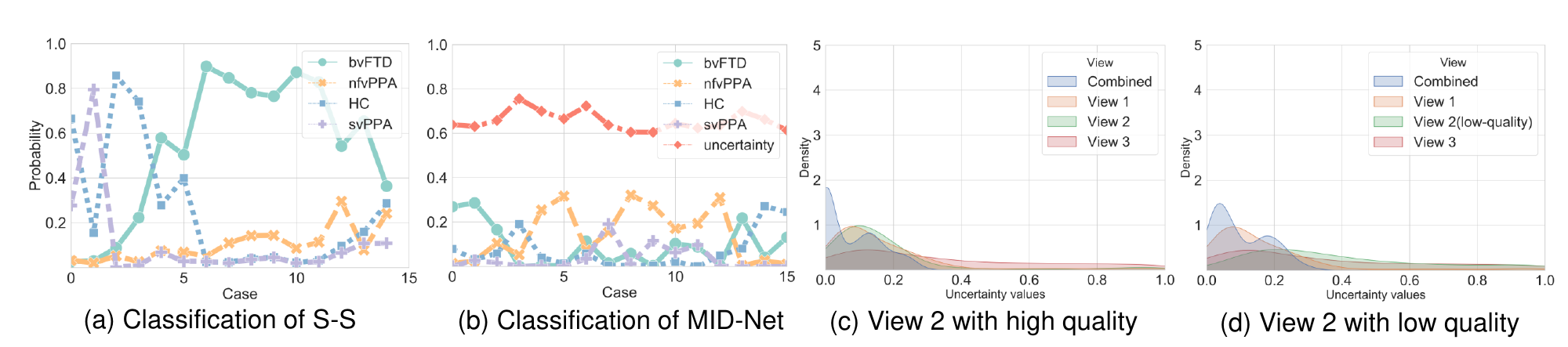}
\caption{\textbf{(a, b)}: The classification probability of the out-of-distribution cases using MID-Net and baseline softmax method.   \textbf{(c)}: High-quality is with the same pre-processing as other views and train set images. \textbf{(d)}: Low-quality means without de-noise processing.}
\end{figure}

\textbf{Uncertainty Estimation.} To further illustrate the role of uncertainty estimation in our model, we visualize the classification results of several out-of-distribution samples from  the public ADHD fMRI image dataset ADHD-200\footnote{\url{http://fcon_1000.projects.nitrc.org/indi/adhd200/}}. As reported in Fig.~\ref{Fig.uncertainty} (a), the baseline model tends to predict very high confidence for the most likely class even if it is completely erroneous. But benefit from the ability to say "I'm not sure.", our MID-Net can detect the decision risk thus provide high uncertainty as shown in \ref{Fig.uncertainty} (b). As shown in the \ref{Fig.uncertainty} (c) and (d), higher uncertainty are generated for low-quality view 2. Meanwhile, the uncertainty of the combined opinoin is higher. These results suggest our model's predictions remain credible even when encountering low-quality data.

\begin{table}[!t]

\caption{Comparison on FTD test set that pre-processed with the same procedure as the training set.}

  \centering
    \begin{tabular}{ccccccccc}
    \toprule
    \multirow{2}[2]*{Method} & \multicolumn{6}{c}{Backbone} \\ \cmidrule(lr){2-7}
    & \multicolumn{3}{c}{Resnet-18} & \multicolumn{3}{c}{VIT}  \\
\cmidrule(lr){2-4} \cmidrule(lr){5-7}     & ACC (\%)       & F1 (\%)    & AUC (\%)  & ACC (\%)       & F1 (\%)    & AUC (\%)\\
    \midrule
    S-S\cite{sarrafclassification}  &82.93 & 80.31 & 96.00 &75.44 &74.52 & 88.57 \\
    S-E\cite{sensoy2018evidential}  &82.39 & 78.38 & 89.72 &75.59 &73.79 & 84.82
\\
    M-S+SF  &93.08 & 92.74 & 98.76  &87.51& 86.66&97.42 \\
    M-E+DS \cite{hantrusted} & 90.99  & 90.81  & 96.24 & 91.30 & 91.04 &96.58  \\
    MID-Net(Ours)   & 91.00    &  90.74 &98.05  & 94.86 & 94.77 & 98.37 \\
    \bottomrule
\label{tab:acc}%
    \end{tabular}%
    
\end{table}%

\textbf{Comparison with baseline methods.}
% \subsection{Results}
As shown in Tab. \ref{tab:acc}, our method reaches satisfactory results. Single-view methods fell to be competitive because of lacking enough brain activity information. With VIT backbone, because of the superiority of IDer, our model far exceeds the accuracy rate of other models by more than 3\%.

\begin{table}[!t]
    \caption{Comparison of FTD classification for low-quality pre-processing.}
  \centering
    \begin{tabular}{ccccccc}
    \toprule
    \multirow{2}[2]*{Method} & \multicolumn{6}{c}{Backbone} \\ \cmidrule(lr){2-7}  & \multicolumn{3}{c}{Resnet-18} & \multicolumn{3}{c}{VIT} \\
\cmidrule(lr){2-4} \cmidrule(lr){5-7}     & ACC (\%)   & F1 (\%)    & AUC (\%)  & ACC (\%)     & F1 (\%)     & AUC (\%)\\
    \midrule

    M-S+SF &76.42$_{\downarrow16.66}$
    &77.26$_{\downarrow15.48}$
    &87.22$_{\downarrow11.54}$
    &78.51$_{\downarrow9.00}$ 
    &76.50$_{\downarrow10.16}$
    &91.80$_{\downarrow5.62}$
 \\
    M-E+DS \cite{hantrusted} &82.79$_{\downarrow8.20}$  
& 81.62 $_{\downarrow9.19}$
&97.30$_{\uparrow1.06}$ 
&86.50$_{\downarrow4.80}$  
&86.15$_{\downarrow4.89}$
&94.02$_{\downarrow2.56}$ 
 \\
    MID-Net(Ours)   &88.74$_{\downarrow2.26}$ &88.38$_{\downarrow2.36}$
    &96.86$_{\downarrow1.19}$ 
&92.19$_{\downarrow2.67}$ 
&92.00$_{\downarrow2.77}$
&98.44$_{\uparrow0.07}$ 
 \\
    \bottomrule
    \end{tabular}%
  \label{tab:accnoisy}%
\end{table}%

\textbf{Comparison of  decision-making capabilities on low-quality fMRI images.} In fMRI pre-processing, manual calibration is often difficult and time-consuming. But the image quality without manual calibration may be poor. In order to test the model classification ability when the image de-noising effect is poor or a view is polluted. In Tab. \ref{tab:acc} and Tab. \ref{tab:accnoisy}, we can see that with both backbones, the performance of our method had only slightly decreased by less than half that of other methods. IDer redistributes beliefs and uncertainty, and combines opinions well, eliminating most of the distractions.

\section{Conclusion}
In this paper, we present MID-Net, a multi-view impartial decision network for FTD diagnosis with uncertainty estimation. Our approach offers a means of estimating the uncertainty of each prediction, which is crucial for providing confidence measurements in FTD diagnosis. To accomplish this, we propose the use of an Impartial Decision Maker (IDer) that can combine opinions impartially and make inferences without incurring computational costs or necessitating changes to the backbone network. As a result, our model can prevent overconfident predictions and accurately estimate the risks associated with its decisions. Our extensive experiments demonstrate that our approach provides reliable and robust uncertainty estimates, which can quantify the decision-making risk of the model. Furthermore, we show that our method can also identify poor quality FTD pre-processing. Moreover, in the diagnosis of FTD, IDer conducts fair unification and reasoning on the evidence of brain activity information collected from different perspectives.
In summary, our MID-Net competes effectively with previous approaches in terms of classification robustness and the reliability of uncertainty estimation. It provides a valuable contribution to the field of FTD diagnosis by offering a reliable and impartial means of decision-making that can accommodate evidence from multiple perspectives.\\

%\bibliographystyle{splncs04}IEEEtran

\bibliographystyle{splncs04}
\bibliography{MFTDnet}

\begin{thebibliography}{10}
\providecommand{\url}[1]{\texttt{#1}}
\providecommand{\urlprefix}{URL }
\providecommand{\doi}[1]{https://doi.org/#1}

\bibitem{bright2019neuroinflammation}
Bright, F., Werry, E.L., Dobson-Stone, C., Piguet, O., Ittner, L.M., Halliday,
  G.M., Hodges, J.R., Kiernan, M.C., Loy, C.T., Kassiou, M., et~al.:
  Neuroinflammation in frontotemporal dementia. Nature Reviews Neurology
  \textbf{15}(9),  540--555 (2019)

\bibitem{dezert2006proportional}
Dezert, J., Smarandache, F.: Proportional conflict redistribution rules for
  information fusion. Advances and applications of DSmT for Information
  Fusion-Collected works  \textbf{2},  3--68 (2006)

\bibitem{do2017plant}
Do, T.B., Nguyen, H.H., Vu, H., Le, T.L., et~al.: Plant identification using
  score-based fusion of multi-organ images. In: 2017 9th International
  conference on knowledge and systems engineering (KSE). pp. 191--196. IEEE
  (2017)

\bibitem{fan2016human}
Fan, L., Li, H., Zhuo, J., Zhang, Y., Wang, J., Chen, L., Yang, Z., Chu, C.,
  Xie, S., Laird, A.R., et~al.: The human brainnetome atlas: a new brain atlas
  based on connectional architecture. Cerebral cortex  \textbf{26}(8),
  3508--3526 (2016)

\bibitem{hantrusted}
Han, Z., Zhang, C., Fu, H., Zhou, J.T.: Trusted multi-view classification. In:
  International Conference on Learning Representations

\bibitem{1255332}
Josang, A., Daniel, M., Vannoorenberghe, P.: Strategies for combining
  conflicting dogmatic beliefs. In: Sixth International Conference of
  Information Fusion, 2003. Proceedings of the. vol.~2, pp. 1133--1140 (2003)

\bibitem{josang2016subjective}
J{\o}sang, A.: Subjective logic, vol.~4. Springer (2016)

\bibitem{mao2019spatio}
Mao, Z., Su, Y., Xu, G., Wang, X., Huang, Y., Yue, W., Sun, L., Xiong, N.:
  Spatio-temporal deep learning method for adhd fmri classification.
  Information Sciences  \textbf{499},  1--11 (2019)

\bibitem{meeter2017imaging}
Meeter, L.H., Kaat, L.D., Rohrer, J.D., Van~Swieten, J.C.: Imaging and fluid
  biomarkers in frontotemporal dementia. Nature Reviews Neurology
  \textbf{13}(7),  406--419 (2017)

\bibitem{pereira2009machine}
Pereira, F., Mitchell, T., Botvinick, M.: Machine learning classifiers and
  fmri: a tutorial overview. Neuroimage  \textbf{45}(1),  S199--S209 (2009)

\bibitem{ramzan2020deep}
Ramzan, F., Khan, M.U.G., Rehmat, A., Iqbal, S., Saba, T., Rehman, A., Mehmood,
  Z.: A deep learning approach for automated diagnosis and multi-class
  classification of alzheimer’s disease stages using resting-state fmri and
  residual neural networks. Journal of medical systems  \textbf{44},  1--16
  (2020)

\bibitem{rashid2020use}
Rashid, M., Singh, H., Goyal, V.: The use of machine learning and deep learning
  algorithms in functional magnetic resonance imaging—a systematic review.
  Expert Systems  \textbf{37}(6),  e12644 (2020)

\bibitem{sarrafclassification}
Sarraf, S., Tofighi, G.: Classification of alzheimer’s disease using fmri
  data and deep learning convolutional neural networks

\bibitem{sensoy2018evidential}
Sensoy, M., Kaplan, L., Kandemir, M.: Evidential deep learning to quantify
  classification uncertainty. Advances in neural information processing systems
   \textbf{31} (2018)

\bibitem{shafer1976mathematical}
Shafer, G.: A mathematical theory of evidence, vol.~42. Princeton university
  press (1976)

\bibitem{10.3389/fninf.2018.00023}
Wen, D., Wei, Z., Zhou, Y., Li, G., Zhang, X., Han, W.: Deep learning methods
  to process fmri data and their application in the diagnosis of cognitive
  impairment: A brief overview and our opinion. Frontiers in Neuroinformatics
  \textbf{12} (2018)

\bibitem{Yan_2016}
Yan, C.G., Wang, X.D., Zuo, X.N., Zang, Y.F.: {DPABI}: Data processing \&
  analysis for (resting-state) brain imaging. Neuroinformatics  \textbf{14}(3),
   339--351 (apr 2016)

\bibitem{zadeh1984review}
Zadeh, L.A.: Review of a mathematical theory of evidence. AI magazine
  \textbf{5}(3),  81--81 (1984)

\bibitem{zhang2018multi}
Zhang, C., Adeli, E., Zhou, T., Chen, X., Shen, D.: Multi-layer multi-view
  classification for alzheimer’s disease diagnosis. In: Proceedings of the
  AAAI Conference on Artificial Intelligence. vol.~32 (2018)

\bibitem{zhang2021survey}
Zhang, X., Yao, L., Wang, X., Monaghan, J., Mcalpine, D., Zhang, Y.: A survey
  on deep learning-based non-invasive brain signals: recent advances and new
  frontiers. Journal of neural engineering  \textbf{18}(3),  031002 (2021)

\end{thebibliography}
\newpage
% \clearpage
\begin{appendix}
\section*{Supplementary materials}

\begin{table}[!htbp]
    \caption{pre-processing procedures using DPABI tool.}
  \centering
    \begin{tabular}{cc}
    \toprule
  \multicolumn{1}{c}{Dataset pre-processing procedures
}   \\ \midrule
Remove first 10 time points\\
Slice Timing\\
Realign\\
Bet\\
T1 Co-register to Fun\\
Nuisance Covariates Regression\\
Affine Regularisation in Segmentation: European\\
Filter(Hz): 0.01$ \sim $ 0.1\\
Normalize + Add mean back\\
Smooth\\

    \bottomrule
    \end{tabular}%
  \label{tab:pre-processing}%
\end{table}%

\begin{figure}[!htbp]
\centering
\label{Fig.compare}
\includegraphics[width = 12cm]{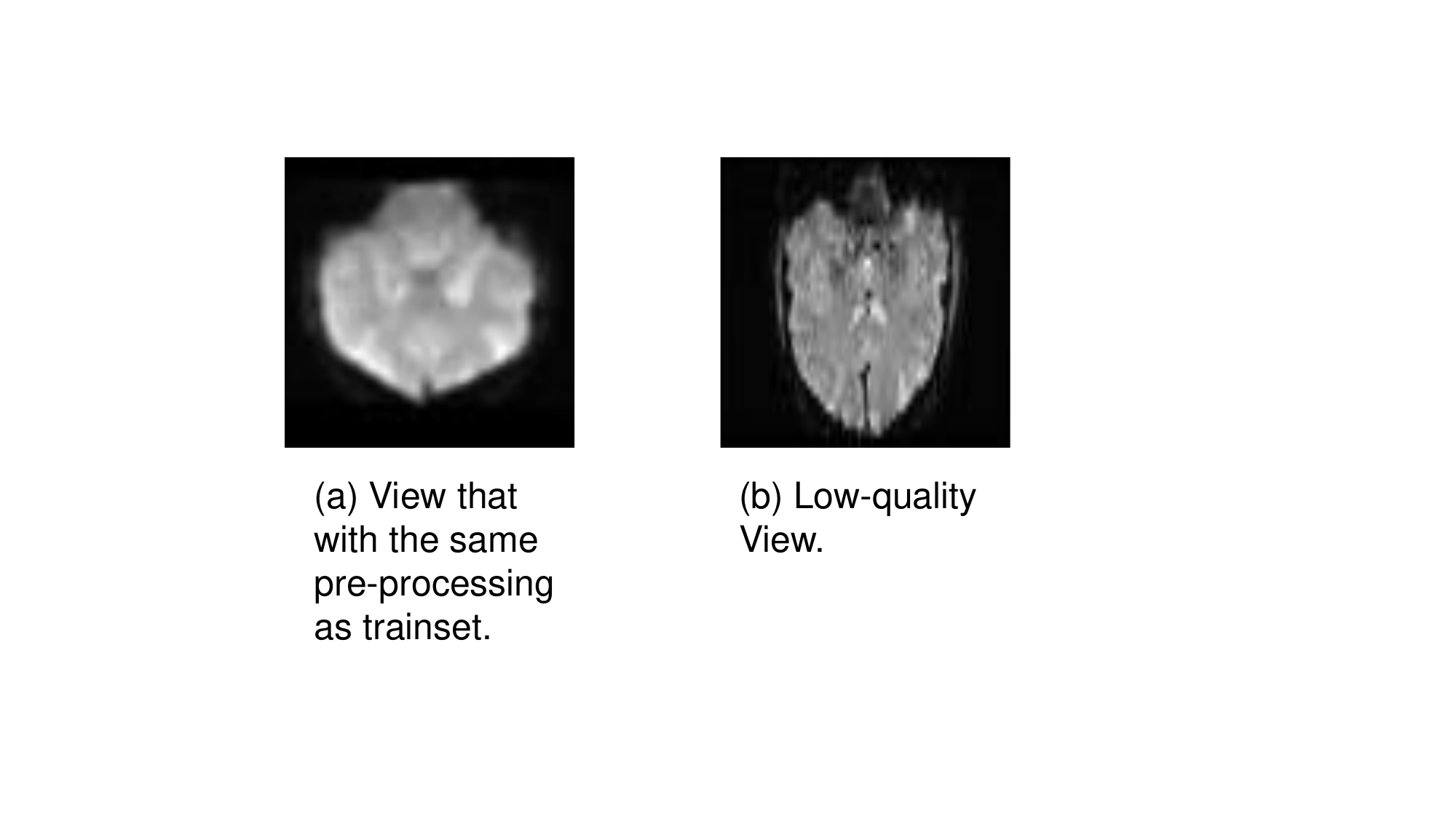}
\caption{Comparison of the same sample with different pre-processing quality. When Fig. (b) is used for input instead of Fig. (a), the output uncertainty for this view increase 34.5\%, the overall uncertainty increase 5\%.}
\end{figure}

\end{appendix}
\end{document}